\documentclass{article}

\usepackage{arxiv}

\pdfoutput=1

\usepackage[utf8]{inputenc}
\usepackage[T1]{fontenc}
\usepackage{hyperref}
\usepackage{url}
\usepackage{booktabs}
\usepackage{amsmath}
\usepackage{amsfonts}
\usepackage{nicefrac}
\usepackage{microtype}
\usepackage{cleveref}
\usepackage{graphicx}
\usepackage{subfigure}
\usepackage[numbers,sort&compress]{natbib}
\usepackage{doi}
\usepackage{authblk}
\usepackage{multirow}
\usepackage{amssymb}
\usepackage{pifont}
\usepackage[ruled,vlined]{algorithm2e}
\usepackage{colortbl}

\title{RAMP: Robust Adaptive Mixed-Precision Quantization for Edge CPU Vision Models}

\date{}

\newbox{\orcid}\sbox{\orcid}{\includegraphics[scale=0.06]{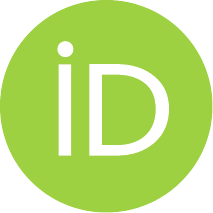}} 
\author[1,2]{%
	\href{https://orcid.org/0009-0002-2054-4446}{\usebox{\orcid}\hspace{1mm}David Población-Criado\thanks{\texttt{david.poblacion@bsc.es}}}%
}
\author[1]{%
	\href{https://orcid.org/0000-0001-6732-5641}{\usebox{\orcid}\hspace{1mm}Dario Garcia-Gasulla}%
}
\author[1]{%
	\href{https://orcid.org/0000-0002-5465-964X}{\usebox{\orcid}\hspace{1mm}Eduardo Quinones}%
}
\affil[1]{Barcelona Supercomputing Center (BSC)}
\affil[2]{Universitat Politècnica de Catalunya - BarcelonaTech (UPC)}

\renewcommand{\shorttitle}{RAMP: Robust Adaptive MPQ for Edge CPU Vision Models}

\renewcommand{\undertitle}{Accepted at the 37th British Machine Vision Conference (BMVC) 2026}
\renewcommand{\headeright}{BMVC 2026}

\hypersetup{
pdftitle={RAMP: Robust Adaptive Mixed-Precision Quantization for Edge CPU Vision Models},
pdfsubject={cs.CV, cs.LG, cs.PF},
pdfauthor={David Población-Criado, Dario Garcia-Gasulla, Eduardo Quinones},
pdfkeywords={Mixed-Precision Quantization, Post-Training Quantization, Solver-Free Optimization, Hardware-Aware Quantization, Edge Vision},
hidelinks
}

\definecolor{failbg}{gray}{0.82}
\newcommand{\clp}[1]{\cellcolor{failbg}#1}

\begin{document}
\maketitle

\begin{abstract}
	Deploying deep learning models on edge CPUs is often bottlenecked by computational and memory constraints. To alleviate this, mixed-precision quantization promises to reduce inference latency while preserving model accuracy. However, quantization affects different layer types in subtle and inconsistent ways, and yet identifying where accuracy loss is minimized and latency reduction is maximized is critical, as the effect accumulates over a full deployment into substantial savings or unacceptable task degradation. Such identification relies on sensitivity metrics, proxies that estimate layer-wise degradation without evaluating the task accuracy of every candidate policy. Nevertheless, widely used metrics fail systematically on modern architectures. We present a systematic empirical study of 13 sensitivity metrics for layer-wise INT8 quantization across four distinctly different neural networks, and validate the resulting policies on two ARM64 platforms. Gradient-based sensitivity methods fail on 4 out of 8 model-hardware configurations and weight-based statistics on 2. In contrast, the Jensen-Shannon Divergence achieves zero catastrophic failures, reliably isolating the layers that cannot be safely quantized. These results complement previous studies by quantifying how much more reliable output-space divergences are than the commonly used gradient- and weight-based criteria. A sensitivity metric alone does not define a policy, and the fixed thresholds typically used for that step are fragile over the highly skewed distributions of modern architectures. We address this with K-Means clustering, achieving near-lossless accuracy and a mean speed-up of $1.81\times$ over the full-precision model. Finally, we reveal that excluding from quantization the layers whose speed-up is negligible, regardless of their sensitivity, can be counterproductive, as it induces computational graph fragmentation and disables operator fusion. Our results yield concrete allocation policies for practitioners and researchers deploying quantized vision models on heterogeneous edge CPUs, without GPU access or gradient computation. Code is available at \url{https://github.com/davidpob99/ramp-mpq}.
\end{abstract}

\keywords{Mixed-Precision Quantization \and Post-Training Quantization \and Solver-Free Optimization \and Hardware-Aware Quantization \and Edge Vision}

\section{Introduction}
\label{sec:intro}
Edge computing has become fundamental to many real-time applications, such as autonomous driving, robotics or mobile image recognition, where the deployment of deep learning models is strictly constrained by latency, energy consumption and memory footprint. These restrictions depend on the complexity of models being executed, which range from Convolutional Neural Networks (CNNs) to Vision Transformers (ViTs). Regardless of the architecture, these models, which are inherently designed for high-end servers, must be adapted to enable their execution on resource-constrained edge devices.

To overcome these limitations, several approaches have been proposed to optimize model deployment. The first involves designing lightweight architectures, often leveraging Neural Architecture Search (NAS) \cite{wang2020apq}. However, this requires a large volume of expensive experiments to find ideal model designs for each specific task, and is typically performed during the model pre-training stage. The second focuses on compressing and accelerating existing, pre-trained models, thereby reducing the overall computational effort. These compression methods range from pruning (removing specific network components) \cite{chengSurveyDeepNeural2024} to quantization (reducing the numerical precision of tensors at the channel, layer, or group level) \cite{deng2020model}. This work focuses on quantization due to its capacity to provide substantial hardware acceleration on edge devices without requiring access to the original training pipeline.

Uniform low-bitwidth quantization yields substantial speed-ups, but it often triggers catastrophic accuracy degradation. Modern architectures mix layer types whose tolerance to precision loss varies by orders of magnitude, so a single network-wide bit-width either sacrifices accuracy on the most sensitive layers or forgoes the available speed-up on the rest. Consequently, mixed-precision quantization (MPQ) has emerged as a robust alternative. Unlike uniform approaches, MPQ frameworks seek an optimal layer-wise bit-width allocation that balances task accuracy with physical hardware constraints, such as inference latency or model size. Existing MPQ approaches are highly diverse, ranging from computationally expensive search-based methods (e.g., Reinforcement Learning (RL)) \cite{wang2019haq, lee-etal-2025-amq} to criterion-based methods using Integer Linear Programming (ILP) \cite{akbulut2026infoq, yao2021hawqv3}. Within these criterion-based methods, bit-width allocation is guided by evaluating layer sensitivity, which is the degree to which the quantization of a specific layer contributes to the overall degradation of the model's performance. Although often effective, these traditional frameworks typically incur significant computational overheads during the policy search phase.

\begin{figure}[h]
	\centering
	\includegraphics[width=0.75\linewidth]{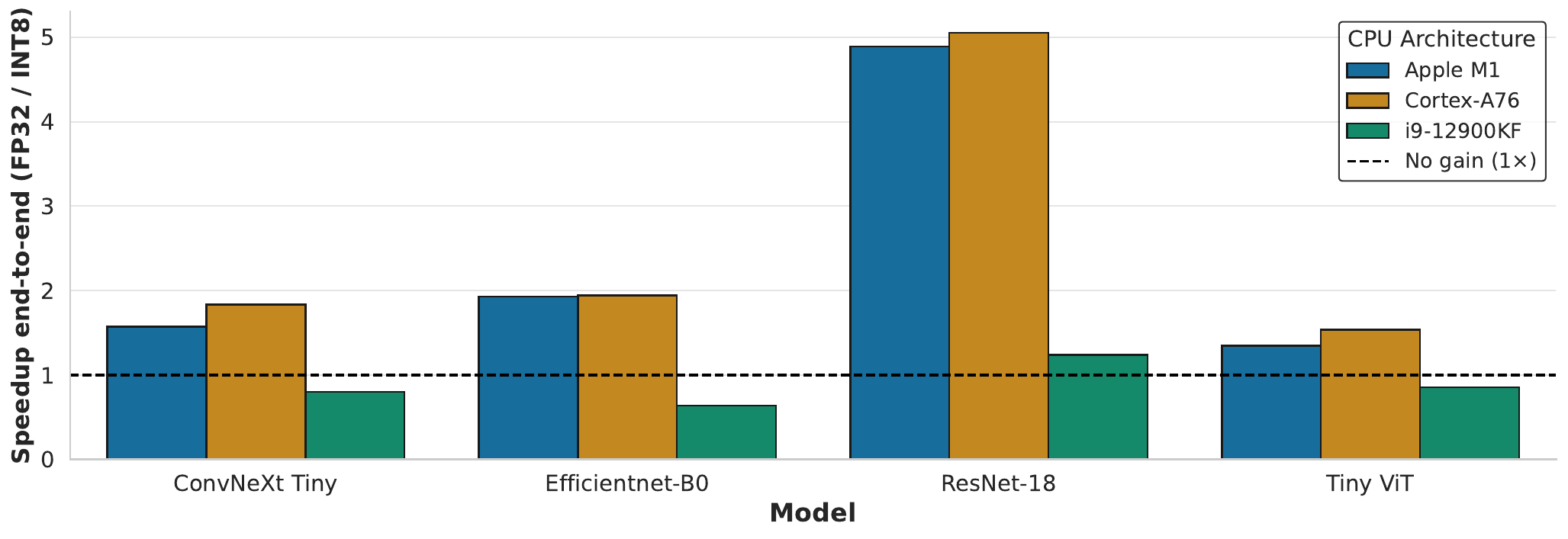}
	\caption{\textbf{End-to-end INT8 inference speed-up relative to FP32} for four vision models and three CPUs under a \emph{single fixed} deployment configuration: ONNX Runtime with the default CPU execution provider, static PTQ and batch size~1. The dashed line marks no gain ($1\times$).}
	\label{fig:speed-up-end-to-end}
\end{figure}

Most existing quantization methods have been optimized for Graphics Processing Units (GPUs). However, many edge deployments rely strictly on general-purpose Central Processing Units (CPUs) due to their power-efficient properties and reduced cost, which is especially relevant in constrained environments. These scenarios exhibit fundamentally different execution characteristics. For instance, edge CPUs typically support a very limited set of native numerical precisions (predominantly 32-bit floating point (FP32) and 8-bit integer (INT8)) at the Instruction Set Architecture (ISA) level, making many sub-byte GPU optimizations inapplicable. Moreover, while the GPU ecosystem is relatively standardized, the CPU landscape is highly heterogeneous, comprising vastly different micro-architectures and instruction sets. Figure~\ref{fig:speed-up-end-to-end} illustrates the consequence. Under a single fixed deployment configuration, the end-to-end benefit of INT8 spans an order of magnitude across different CPUs, from a $5\times$ acceleration to a net slowdown. This variance underscores the importance of accounting for the target execution stack when designing quantization policies.

To address these challenges, this work first conducts a systematic evaluation of existing sensitivity proxies and hardware profiling methods. We find that traditional metrics often fail to predict accuracy degradation in modern architectures. Additionally, standard latency modelling treats hardware awareness as a set of layer-level constraints, overlooking critical global-level overheads like computational graph fragmentation. We show that this fragmentation disables operator fusion and increases execution costs, which can lead to suboptimal deployment. Furthermore, we observe that common thresholding methods rely on fixed percentiles, which are fragile as they fail to account for the diverse sensitivity distributions of different architectures, while search-based methods are computationally expensive and unsuitable for resource-constrained CPUs. To resolve these issues, we propose a solver-free, criterion-based clustering method that uses K-Means to adaptively identify sensitivity breakpoints, optimizing the accuracy-latency trade-off on CPUs. Our primary contributions are summarized as follows:

\begin{itemize}
	\item \textbf{Systematic Reassessment of Sensitivity Proxies:} We show that Jensen-Shannon Divergence (JSD) is a better proxy for quantization degradation than gradient-based metrics, successfully preventing accuracy collapse in diverse architectures.
	\item \textbf{Revisiting Latency Modelling and Graph Fragmentation:} We reveal that strictly enforcing layer-level constraints induces computational graph fragmentation and disables operator fusion, which outweighs the benefits of individual operator acceleration.
	\item \textbf{RAMP (Robust Adaptive Mixed-Precision Quantization):} We introduce a solver-free method that applies 1D K-Means clustering to the JSD landscape to automatically generate adaptive and near-lossless quantization policies.
	
\end{itemize}

\section{Background and Related Work}
\label{sec:background}
This section reviews the fundamentals of mixed-precision quantization and existing sensitivity proxies, followed by an overview of the hardware constraints that motivate our method.

\subsection{Mixed-Precision Quantization}
Fixed post-training quantization (PTQ) relies on applying a uniform bit-width across the entire network. However, this is inherently suboptimal \cite{dong2023emq}, as layers differ in their sensitivity to precision loss. Mixed-precision quantization (MPQ) is the process of assigning different numerical precisions to specific layers, channels, or tensors within a neural network. The primary objective is to maintain the most sensitive layers at a high precision while aggressively compressing those that contribute the least to overall model performance. This optimization problem grows exponentially with model size: given a candidate precision set $B$ and $L$ layers, the search space encompasses $\mathcal{O}(|B|^L)$ possible combinations. To explore this huge space, MPQ methods are generally classified into two categories based on their exploration strategy: \textit{search-based} and \textit{criterion-based}. In this paper, we focus on the latter, leveraging proxy metrics to quantify the degradation each individual layer introduces to the model.

\paragraph{Search-based.} These approaches empirically explore combinations within the design space using techniques such as Reinforcement Learning (RL) \cite{wang2019haq, lou2020autoq}, genetic algorithms \cite{lee-etal-2025-amq}, or joint architecture search \cite{wang2020apq, wu2018dnas, cai2020edmips, guo2020single, bai2021batchquant}. Consequently, their primary drawback is not only the severe computational overhead, often requiring days or weeks of compute to converge, but also the strict requirement for full access to the original training pipeline and entire dataset. In contrast, criterion-based methods guide allocation analytically rather than through empirical search. This avoids the expensive policy generation phase, making them vastly more efficient and practical for post-training edge deployment.

\paragraph{Criterion-based.} These methods use computationally inexpensive proxy metrics to estimate the sensitivity of each layer to quantization. Sensitivity is defined as the extent to which quantizing a specific layer contributes to accuracy degradation or an increase in the loss function. In the literature, sensitivity proxies are generally divided into three main categories. First, \textit{Second-Order (Curvature-Based) Metrics} estimate sensitivity based on the local curvature of the loss landscape, typically captured by the Hessian matrix via Taylor expansions (e.g., HAWQ-V2 \cite{dongHAWQV2HessianAware2020}). Second, \textit{Information-Theoretic Metrics} are used in two different ways. Some methods quantify the information lost between the full-precision and the quantized model, typically via Kullback-Leibler (KL) divergence \cite{ranjan2025mixqsam, liu2026sigmaquant}. Others score the information content of the model itself without comparing the two: Sun et al \cite{sun2022entropydrivenmpq} derive the entropy of the last feature map, whereas Qin et al \cite{qin2025mixed} measure the Shannon entropy of each layer's activations and assign bit-widths against fixed percentiles of the resulting distribution. Finally, \textit{Signal Distortion Metrics} rely on quantifying the local discrepancy between full-precision weights and their low-precision counterparts, employing standard distance measurements such as Signal-to-Quantization Noise Ratio (SQNR) \cite{pandey2023practical} or Mean Squared Error (MSE) \cite{choukroun2019lowbit}.

Given a specific sensitivity metric, these approaches formulate an ILP \cite{akbulut2026infoq, yao2021hawqv3} or Quadratic Programming (QP) \cite{ranjan2025mixqsam, deng2025mixedprecision} problem to minimize it subject to a set of constraints, which are primarily hardware-oriented, such as model size or inference latency \cite{wang2019haq,yao2021hawqv3}. This formulation bridges two orthogonal components: sensitivity, acting as a proxy for task degradation, and the physical constraints of the target hardware. However, because ILP solvers introduce significant computational overhead, many practical frameworks resort to rigid, rule-based heuristics, such as percentile thresholding \cite{qin2025mixed, chen2025cmpq}. While faster, these static thresholds fail to account for the highly skewed sensitivity distributions inherent to modern network architectures, directly motivating the need for more adaptive grouping mechanisms.

\subsection{Target Hardware}
Most MPQ methods target GPUs \cite{dongHAWQV2HessianAware2020, yao2021hawqv3, akbulut2026infoq}, leaving CPU deployment under-explored. Unlike GPUs, edge CPUs exhibit high micro-architectural heterogeneity that complicates latency estimation. \textit{First}, the physical implementation of instruction sets varies significantly across architectures (e.g., ARM64, x86-64), which can lead to counter-intuitive execution profiles where higher-precision operations are executed faster than lower-precision ones. \textit{Second}, CPU performance is heavily bottlenecked by localized memory hierarchies (L1/L2 caches); the overhead of data movement often dictates the overall execution latency, making cache utilization as critical as arithmetic intensity. To address this gap in the literature, this study optimizes MPQ for general-purpose CPUs, centring its empirical scope to ARM64: it is the main edge deployment target and exposes INT8 uniformly at the ISA level, so the latency variance we observe is attributable to micro-architecture rather than to instruction availability.

\section{Problem Formulation}
Let a Deep Neural Network (DNN) be defined by a sequence of $L$ layers parameterized by full-precision weights $W$. In MPQ, we define the set of available bit-widths as $B = \{b_0, ..., b_{n-1}\}$. For each layer $l \in \{1, ..., L\}$, we assign a specific bit-width precision to its weights ($w_l$) and activations ($a_l$). To mitigate hardware deployment overhead and simplify the optimization space, we enforce a coupled quantization strategy where weights and activations within the same layer must share the same bit-width precision ($w_l = a_l \in B$). Consequently, the precision configuration of each layer is defined by a single scalar $p_l \in B$. In the simplest MPQ scenario, deciding whether to quantize a layer to a target low-precision bit-width or keep it in full precision, we have $|B| = 2$. Under this coupled constraint, the total search space is reduced to $|\mathcal{A}| = |B|^L = 2^L$ possible configurations.

The primary objective of MPQ is to discover the optimal bit-width configuration within the search space $\mathcal{A}$ that minimizes the task loss (i.e., accuracy degradation) subject to a resource constraint. However, directly evaluating the true task loss across all candidates is computationally prohibitive. To circumvent this, a sensitivity metric $M$ is typically employed as a proxy to quantify the local degradation between the full-precision baseline and the quantized layer. Formally, the sensitivity penalty $\Omega_l$ for a given layer $l$ is defined in Equation~\ref{eq:sensitivity}, where $f_l$ and $\hat{f}_l$ denote the full-precision and quantized representations (or outputs) of layer $l$, respectively.

\begin{equation}
	\label{eq:sensitivity}
	\Omega_l = M(f_l, \hat{f}_l)
\end{equation}
Consequently, the optimization problem can be formulated as a discrete subset selection task as in Equation~\ref{eq:min_sensitivity}.
\begin{equation}
	\label{eq:min_sensitivity}
	\min_{\mathcal{A}} \sum_{l=1}^{L} \Omega_l \quad \text{s.t.} \quad Cost(\mathcal{A}) \leq C
\end{equation}
Typical $Cost$ functions often account for model memory footprint or theoretical computational complexity (e.g., bit-operations). While such constraints are traditionally optimized using ILP, our solver-free approach bypasses these explicit restrictions by leveraging clustering algorithms to derive the quantization policy.

Therefore, defining a complete MPQ criterion-based method requires two fundamental components: (i) a sensitivity metric ($M$) to quantify the impact of quantization and (ii) a search method to determine the optimal bit-width allocation.

\section{Mixed-Precision Quantization}
\label{sec:mpq}
This section establishes the sensitivity metric and the search policy. We first define an experimental setup across diverse architectures. Next, we analyse multiple proxies to identify the most effective metric for ranking sensitivity degradation. Finally, we introduce a distribution-adaptive thresholding technique based on K-Means clustering, which will serve as the core search engine for our subsequent quantization method.

\subsection{Experimental Setup}
We conduct a sensitivity analysis focused on accuracy preservation across diverse model architectures: ResNet-18 \cite{he2016deep}, EfficientNet-B0 \cite{tan2019efficientnet}, ConvNeXt-Tiny \cite{liuConvNet2020s2022} and TinyViT \cite{wu2022tinyvit}. We selected these models to cover a wide architectural spectrum, ranging from classical CNNs to modern ViTs. We evaluate on SuSy-Dataset \cite{bernabeu-perez2026present}, a synthetic image classification dataset, to assess the impact of quantization beyond standard classification tasks. Deepfake detection relies on subtle high-frequency artifacts \cite{frank2020leveraging, qian2020thinking} that may be more susceptible to quantization-induced distortion than the coarse semantic features used in ImageNet classification, making this task a challenging testbed for our accuracy sensitivity experiments.

\subsection{Sensitivity Analysis}
We first evaluate the capacity of various sensitivity metrics to predict empirical accuracy degradation. To comprehensively assess these predictive proxies for quantization robustness, we select 13 diverse metrics from the literature and categorize them into three distinct representational spaces:
\begin{itemize}
	\item \textbf{Parameter-Space Proxies.} These metrics assess local tensor degradation based on structural properties (e.g., parameter count), statistical dispersion (standard deviation (STD), coefficient of variation of channel ranges, dynamic range ratios) and quantization noise heuristics (Signal-to-Noise Ratio (SNR) and Estimated Signal-to-Quantization-Noise Ratio (SQNR)).
	\item \textbf{Gradient-Space Proxies.} Represented by the state-of-the-art Hessian trace approximation (HAWQ-V2  score \cite{dongHAWQV2HessianAware2020}), this category captures the local curvature of the loss landscape to estimate how weight perturbations affect the overall loss.
	\item \textbf{Output-Space Proxies.} These proxies quantify the end-to-end semantic degradation via distance metrics (MSE, NRMSE, MAE, Cosine Dissimilarity) and information-theoretic divergences (Kullback-Leibler (KL) and Jensen-Shannon Divergence (JSD)) computed directly on the final classification logits.
\end{itemize}

\paragraph{One-At-a-Time Measurement.} To evaluate each sensitivity metric, we employ a One-At-a-Time (OAT) analysis. First, we extract the reference weights and output logits from the full-precision baseline model. Subsequently, the network is independently quantized $L$ times, quantizing only a single layer $l$ per iteration while retaining the rest in full precision. The sensitivity metrics are then computed by comparing the full-precision baseline against each of the $L$ resulting models. This approach ensures that we capture the global impact of a single-layer perturbation on the network's overall dynamics and final output logits, rather than measuring isolated local errors for output-space proxies \cite{akbulut2026infoq}. Nevertheless, this method ignores inter-layer dependencies and error accumulation under joint quantization, a limitation shared by criterion-based MPQ methods \cite{dongHAWQV2HessianAware2020,yao2021hawqv3}. What OAT does provide is a ranking of individually sensitive layers, which Section~\ref{sec:ramp} shows to be sufficient to build joint policies that avoid accuracy collapse.

\paragraph{Evaluation Criteria.} To determine which of these metrics best explains accuracy degradation, we establish a multi-phase evaluation protocol. First, we compute pairwise Spearman correlations \cite{spearman1904proof} across all evaluated proxies to identify redundancies. Second, we correlate the remaining metrics directly against the empirical accuracy degradation, calculated via the OAT approach. We use this correlation as a sanity check rather than as a selection criterion, since output-space proxies are computed on the same logits whose argmax defines accuracy and are therefore favoured by construction. However, in MPQ, identifying the exact magnitude of the accuracy drop or its global correlation is secondary to correctly ranking and retrieving the most vulnerable layers. Therefore, as our primary evaluation criterion, we introduce a Recall@Top-$N\%$ metric. For a given percentile threshold $N\%$, we define the number of target layers as $k = \lfloor L \times \frac{N}{100} \rfloor$. We construct two subsets of cardinality $k$: $\mathcal{L}_{\text{true}}$, containing the top-$k$ most sensitive layers according to the ground-truth accuracy drop $y$ and $\mathcal{L}_{\text{proxy}}$, containing the top-$k$ layers predicted by a given proxy metric $x \in X$. The evaluation metric is defined as in Equation~\ref{eq:topk}.
\begin{equation}
	\label{eq:topk}
	\text{Recall@Top-}k = \frac{|\mathcal{L}_{\text{true}} \cap \mathcal{L}_{\text{proxy}}|}{k}
\end{equation}
Given that $|\mathcal{L}_{\text{true}}| = |\mathcal{L}_{\text{proxy}}| = k$, precision and recall are mathematically equivalent in this context. This bounds the score to $[0, 1]$, where $1.0$ represents a perfect topological match for the most critical layers.

\paragraph{Collapse Criterion.} At the deployment level we also need a binary notion of policy failure, used throughout the paper to report failure counts. We define a \textit{collapse} as a Top-1 accuracy falling more than $15$ percentage points below the FP32 baseline: a deliberately conservative bar, flagging models rendered unusable for their task rather than borderline regressions. Its precise value is not critical, as no observed configuration falls in the $8$--$15$ point range and any threshold within that interval yields identical failure counts.

\paragraph{Empirical Results.} We first compute pairwise Spearman correlations across all evaluated proxies to identify redundancies. Information-theoretic metrics (KL, JSD) and distance-based metrics (MSE, MAE, cosine dissimilarity) are highly inter-correlated ($r > 0.9$), forming a single redundant group. The Hessian-based HAWQ-V2 \cite{dongHAWQV2HessianAware2020} shows moderate correlation with this group ($r \simeq 0.6$), while SNR exhibits an inverse relationship. To assess predictive power, we correlate each metric directly against the per-layer accuracy drop. The results vary across architectures: ConvNeXt-Tiny yields the most significant correlations ($p < 0.05$), followed by ResNet-18 and EfficientNet-B0, whereas no metric reaches statistical significance for TinyViT. Therefore, having established the ground-truth accuracy degradation ($y$) via the OAT approach, we evaluate the proxies strictly on their retrieval performance using the Recall@Top-$N\%$ agreement.

\begin{table*}[!h]
	\centering
	\resizebox{\textwidth}{!}{%
		\begin{tabular}{l ccc ccc ccc ccc c}
			\toprule
			& \multicolumn{3}{c}{\textbf{ResNet-18}} 
			& \multicolumn{3}{c}{\textbf{ConvNeXt-Tiny}} 
			& \multicolumn{3}{c}{\textbf{EfficientNet-B0}} 
			& \multicolumn{3}{c}{\textbf{TinyViT}} 
			& \\
			\cmidrule(lr){2-4} \cmidrule(lr){5-7} \cmidrule(lr){8-10} \cmidrule(lr){11-13}
			\textbf{Metric} 
			& @10\% & @20\% & @30\% 
			& @10\% & @20\% & @30\% 
			& @10\% & @20\% & @30\% 
			& @10\% & @20\% & @30\% 
			& \textbf{Avg.} \\
			& \scriptsize{(k=2)} & \scriptsize{(k=4)} & \scriptsize{(k=6)}
			& \scriptsize{(k=7)} & \scriptsize{(k=15)} & \scriptsize{(k=23)}
			& \scriptsize{(k=8)} & \scriptsize{(k=16)} & \scriptsize{(k=24)}
			& \scriptsize{(k=6)} & \scriptsize{(k=13)} & \scriptsize{(k=20)}
			& \\
			\midrule
			JSD           & \textbf{.500} & \textbf{.750} & \textbf{.667} & \textbf{1.000} & \underline{.800} & \underline{.783} & \textbf{.875} & \underline{.812} & \textbf{.625} & \textbf{.833} & \textbf{.846} & \underline{.700} & \textbf{.766} \\
			KL            & \textbf{.500} & \textbf{.750} & \textbf{.667} & \textbf{1.000} & \underline{.800} & \underline{.783} & \textbf{.875} & \underline{.812} & \textbf{.625} & \textbf{.833} & \textbf{.846} & \underline{.700} & \textbf{.766} \\
			MSE           & \textbf{.500} & \textbf{.750} & \underline{.500} & \underline{.714} & \underline{.800} & .696 & \underline{.750} & \underline{.812} & \textbf{.625} & \textbf{.833} & \underline{.769} & \underline{.700} & .704 \\
			NRMSE         & \textbf{.500} & \textbf{.750} & \underline{.500} & \underline{.714} & \underline{.800} & .696 & \underline{.750} & \underline{.812} & \textbf{.625} & \textbf{.833} & \underline{.769} & \underline{.700} & .704 \\
			Cos.\ Dis.    & \textbf{.500} & \textbf{.750} & \underline{.500} & \underline{.714} & \textbf{.867} & \textbf{.826} & \underline{.750} & \textbf{.875} & \textbf{.625} & \textbf{.833} & \underline{.769} & \underline{.700} & \underline{.726} \\
			MAE           & \textbf{.500} & \textbf{.750} & \underline{.500} & \underline{.714} & \underline{.800} & .696 & \underline{.750} & \underline{.812} & \textbf{.625} & \textbf{.833} & \underline{.769} & .650 & .700 \\
			HAWQ-V2 \cite{dongHAWQV2HessianAware2020} & \textbf{.500} & \underline{.500} & .333 & .571 & .667 & .652 & .375 & .562 & .500 & .333 & .462 & .650 & .509 \\
			STD           & \textbf{.500} & \underline{.500} & \underline{.500} & .143 & .333 & .435 & .500 & .688 & .500 & \underline{.667} & .538 & .650 & .496 \\
			Ch.\ Range CV & \textbf{.500} & .250 & .333 & .429 & .333 & .348 & .625 & .562 & \underline{.583} & .333 & .538 & \textbf{.750} & .465 \\
			Range Ratio   & \textbf{.500} & .250 & .167 & .286 & .267 & .348 & .125 & .312 & .417 & .167 & .385 & \underline{.700} & .327 \\
			Est.\ SQNR    & \textbf{.500} & .250 & \textbf{.667} & .286 & .467 & .435 & .000 & .062 & .167 & .167 & .077 & .150 & .269 \\
			SNR           & .000 & .000 & .333 & .000 & .000 & .000 & .000 & .000 & .208 & .000 & .000 & .000 & .045 \\
			N. Parameters & .000 & .000 & .000 & .000 & .000 & .043 & .000 & .125 & .208 & .000 & .000 & .000 & .031 \\
			\bottomrule
		\end{tabular}%
	}
	\caption{Recall @ Top-$N\%$ of most sensitive layers across architectures. $k$ denotes the number of retrieved layers at each threshold. \textbf{Bold}: best. \underline{Underline}: second best.}
	\label{tab:recall_topk}
\end{table*}

Table \ref{tab:recall_topk} summarizes the Recall@Top-$N\%$ across architectures, yielding four key insights. \textit{First}, output-space metrics (JSD, KL) significantly outperform weight- and gradient-based heuristics, averaging $>0.76$ mean recall. They are closely followed by cosine dissimilarity ($0.726$ mean recall), indicating that measuring directional shifts in activations is also a highly effective proxy for sensitivity. We select JSD for our framework, which compares the full-precision and quantized output distributions $P$ and $Q$ against their mixture $M = \frac{1}{2}(P+Q)$ as $\text{JSD}(P \parallel Q) = \frac{1}{2}D_{\text{KL}}(P \parallel M) + \frac{1}{2}D_{\text{KL}}(Q \parallel M)$. KL achieves identical recall for every architecture and threshold but we nonetheless prefer JSD, which unlike KL is symmetric and bounded to $[0,1]$ \cite{lin1991jsd}. \textit{Second}, the Hessian-based state-of-the-art (HAWQ-V2) struggles on modern architectures (e.g., scoring $0.333$ on TinyViT Top-10\%, compared to JSD's $0.833$). \textit{Third}, Recall@Top-$N\%$ highlights JSD's practical efficacy: while hybrid architectures like TinyViT disrupt global ranking correlations (Spearman), JSD remains a highly precise filter for catastrophically vulnerable layers ($0.833$ Top-10\% Recall). \textit{Finally}, local noise heuristics (SNR, Est. SQNR) perform poorly ($0.045$ and $0.269$ mean, respectively), confirming that tensor-level degradation is a weak proxy for end-to-end task failure.

\subsection{Distribution-Adaptive Thresholding}
While a sensitivity metric like JSD accurately quantifies the expected degradation from quantizing a specific layer $l$, it is necessary to establish a policy that selects layers based on this metric. To solve the constrained optimization problem defined in Equation~\ref{eq:min_sensitivity}, traditional algorithms typically rely on ILPs \cite{akbulut2026infoq, yao2021hawqv3} or, in recent years, RL \cite{lou2020autoq, wang2020apq}, which are computationally expensive. Another approach is to use rigid heuristic methods, such as percentile-based selection, which lack robustness against outliers.

To establish an automated method that accounts for the disparity in the sensitivity metric distribution (without assuming normality), we introduce a distribution-adaptive thresholding mechanism based on K-Means clustering \cite{Lloyd1982}. Let $\Omega = \{\Omega_1, \Omega_2, \dots, \Omega_L\}$ be the continuous sensitivity scores extracted via our OAT profiling for all $L$ layers. Our objective is to partition $\Omega$ into $K$ discrete groups $C = \{c_1, \dots, c_K\}$, where $K$ determines the granularity of the quantization policy (e.g., $K=5$ for fine-grained separation). We solve this by minimizing the within-cluster variance. Formally, the algorithm finds the optimal partition $C$ that minimizes the objective in Equation~\ref{eq:kmeans}, where $\mu_k$ represents the geometric centroid (mean sensitivity) of cluster $c_k$. By optimizing this objective, the K-Means algorithm naturally adapts to the intrinsic skewness of the architecture's distribution. Rather than enforcing an arbitrary layer count per group, it seamlessly aggregates the massive floor of structurally robust layers into a single low-sensitivity cluster, while precisely isolating topological outliers (i.e., highly sensitive layers) into smaller, high-sensitivity micro-clusters.

\begin{equation}
	\label{eq:kmeans}
	\arg\min_{C} \sum_{k=1}^{K} \sum_{\Omega_l \in c_k} ||\Omega_l - \mu_k||^2
\end{equation}

Finally, the quantization bit-width is mapped inversely to the cluster centroids. The layer cluster with the lowest centroid $\mu_k$ is safely mapped to INT8, maximizing hardware acceleration, while the clusters containing the extreme topological outliers are protected in full precision (FP32) to strictly preserve the network's semantic integrity.

\begin{figure}[htbp]
	\centering
	\subfigure[Topology vs. JSD Sensitivity Spectrum]{
		\includegraphics[width=0.43\textwidth]{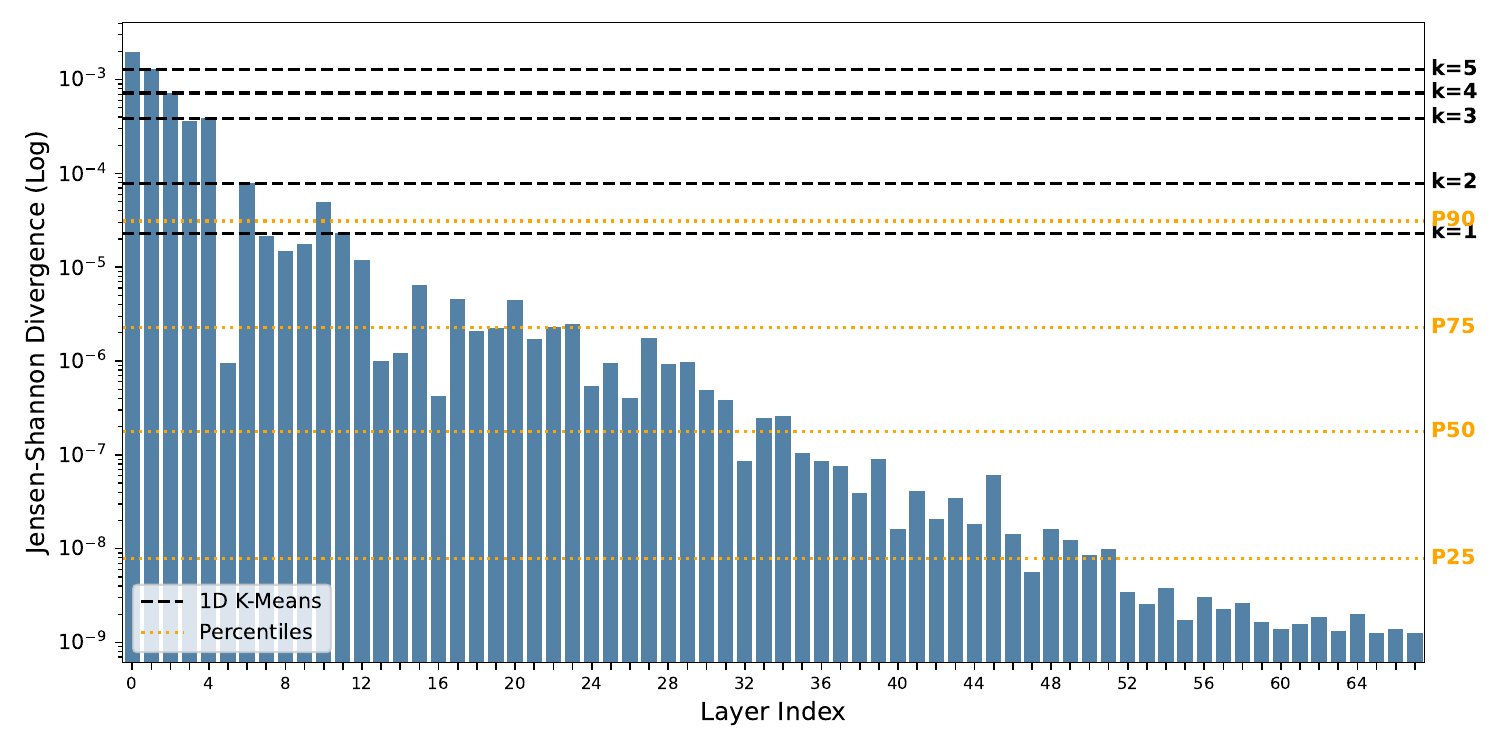}
	}
	\hfill
	\subfigure[Silhouette Clustering Quality]{
		\includegraphics[width=0.52\textwidth]{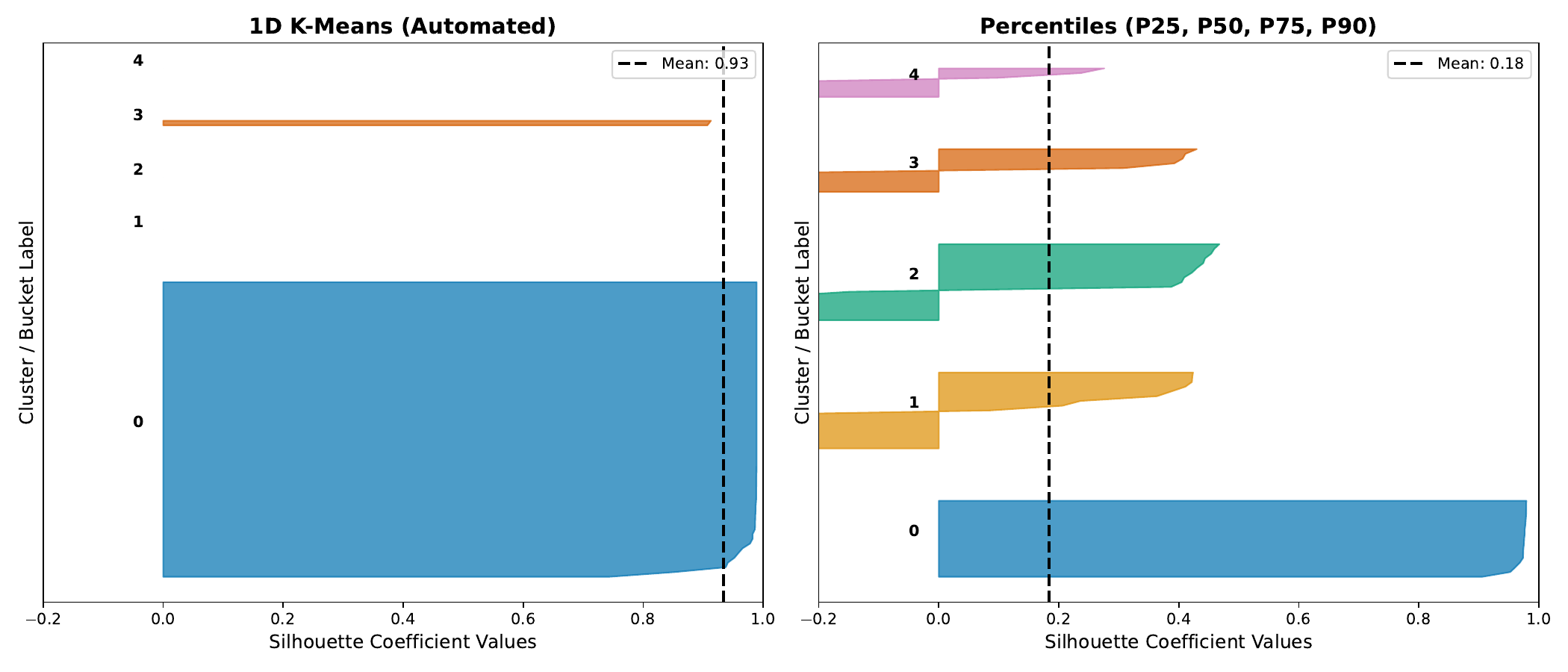}
	}
	\caption{Ablation study of policy thresholding on TinyViT. (a) Ordered layer-wise JSD distribution overlaid with automated 1D K-Means (black) and static percentile (orange) boundaries. (b) Corresponding Silhouette analysis highlighting the topological fragmentation (negative values) caused by fixed percentile quotas versus the highly cohesive clustering achieved by our solver-free approach.}
	\label{fig:ablation_kmeans}
\end{figure}

\paragraph{Clustering Evaluation.} To evaluate the 1D K-Means algorithm ($K=5$), we compared it against a heuristic selection using fixed percentiles ($P=\{25, 50, 75, 90\}$). Figure~\ref{fig:ablation_kmeans} displays the results for TinyViT. Figure~\ref{fig:ablation_kmeans}(a) illustrates the clustering thresholds generated by each method over the JSD (in log scale), demonstrating that K-Means correctly isolates the vast majority of low-sensitivity layers right from the first threshold, whereas the percentile method fails to do so until the 90th percentile. Furthermore, Figure~\ref{fig:ablation_kmeans}(b) presents the Silhouette plot \cite{rousseeuw1987silhouettes}, confirming that K-Means clusters exhibit a superior inter/intra-cluster variance ratio. In contrast, the percentile baseline yields negative silhouette values, mathematically proving that rigid quotas arbitrarily fragment groups of structurally identical layers. The remaining evaluated models suffer from this same behaviour due to the highly skewed nature of modern sensitivity distributions; ResNet presents the smoothest distribution, followed by ConvNeXt, EfficientNet and TinyViT. This progression strictly aligns with the quantization difficulty observed in our previous experiments, demonstrating that fixed percentiles trigger accuracy collapse by blindly quantizing topological outliers, whereas our K-Means approach automatically preserves model semantics.

\section{Hardware Impact}
\label{sec:methodology_hardware}
While MPQ is often used to balance model accuracy and inference latency, assuming that lower precision automatically speeds up execution is often wrong in practice. Even though INT8 theoretically offers better computational speed and uses less memory bandwidth than standard FP32, these benefits heavily depend on the specific hardware architecture, instruction sets and the computational workload of each layer.

To test the impact of hardware, we profile the models of Section~\ref{sec:mpq} on three CPUs from different vendors: an Apple M1, a Raspberry Pi 5 (Cortex-A76) and an Intel Core i9-12900KF. The first two are ARM64 and constitute our main evaluation setting, while the third is x86-64 and serves to characterize cross-ISA variability. All experiments use ONNX Runtime \cite{onnxruntime}, which supports node-level MPQ and is widely used in production, with the default CPU execution provider and only FP32 and INT8 precisions, since both are natively supported by these ISAs.

\begin{figure}[h]
	\centering
	\subfigure[Apple M1]{
		\includegraphics[width=0.48\textwidth]{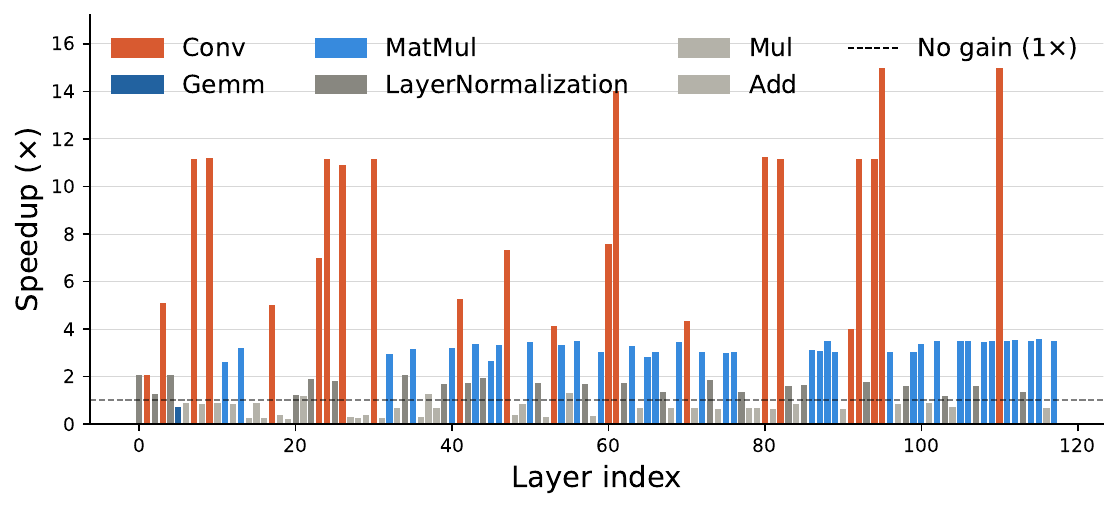}
	}
	\hfill
	\subfigure[Cortex-A76]{
		\includegraphics[width=0.48\textwidth]{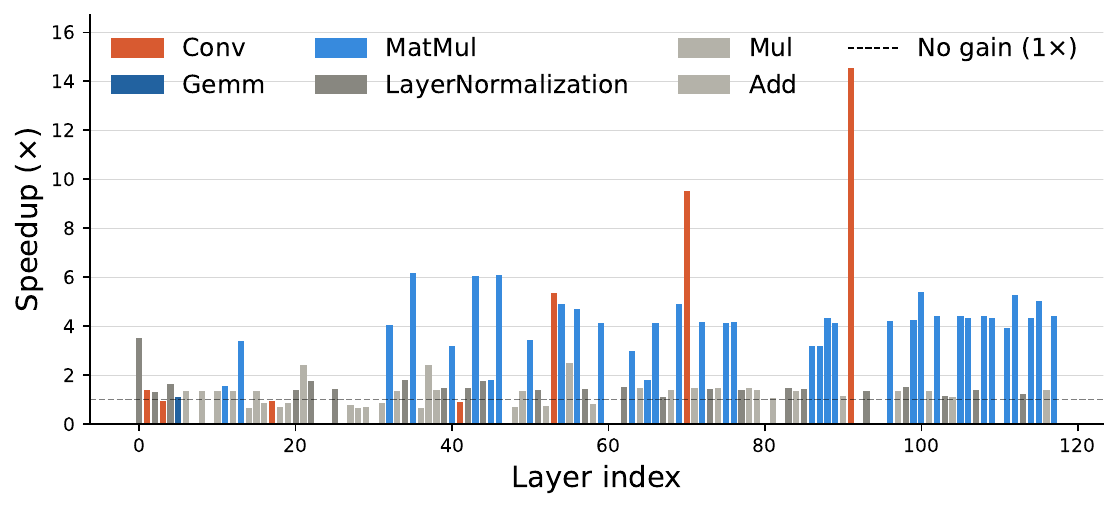}
	}
	\caption{Per layer INT8 speed-up of ConvNeXt-Tiny in different CPUs. Convolutional layers (orange) achieve up to $15\times$ speed-up while normalization and element-wise operations (gray) show minimal gains, motivating a layer-wise quantization policy.}
	\label{fig:speed-up_layer_convnext-tiny}
\end{figure}

As anticipated, the speed-up achieved by INT8 quantization is not uniform across CPU architectures. Figure~\ref{fig:speed-up_layer_convnext-tiny} illustrates the per-layer speed-up for ConvNeXt-Tiny on both platforms, measured as the difference in ONNX Runtime profiling time between FP32 and INT8 execution. Despite the identical model and runtime configuration, the two platforms behave markedly differently: convolutional layers reach $6\times$ to $15\times$ on the Apple M1, whereas on the Cortex-A76 only a minority exceed $6\times$. All tested models present similar behaviours.

This disparity becomes more pronounced when comparing platforms with different ISAs. On x86-64, the default ONNX Runtime CPU provider not only fails to accelerate inference under INT8, but in several cases degrades latency relative to FP32 execution. Switching to OpenVINO \cite{openvino}, Intel's preferred execution backend for deep neural networks, recovers positive speed-ups, though these remain modest, consistently below $1.4\times$.

Several factors contribute to this architecture-dependent behaviour. \textit{First}, ISA differences directly affect integer arithmetic throughput: in contrast to ARM64 platforms with tightly integrated NEON/AMX low-precision execution paths, x86-64 INT8 performance is more heterogeneous, as acceleration depends on the presence of AVX2/AVX512 VNNI or AMX instructions and on backend-specific kernel selection mechanisms (e.g., oneDNN \cite{onednn} or FBGEMM \cite{fbgemm}). \textit{Second}, differences in L1/L2 cache capacity and memory bandwidth affect weight loading latency, which dominates runtime for small tensors \cite{williams2009roofline}. \textit{Third}, and critically, converting weights and activations to INT8 does not guarantee improved latency, because quantization and dequantization operations introduce additional data movement overhead that in some cases surpasses the performance benefits of using lower precision, an effect that is particularly pronounced for layers with low arithmetic intensity, such as small depthwise convolutions, where the cost of type conversion exceeds the savings from reduced computation \cite{Yao2022ZeroQuant, shalby2026dqt}.

These observations have a direct consequence for MPQ policy design. Under the additive cost model prevalent in the literature, where end-to-end latency is treated as the sum of independent layer costs, they imply that a layer yielding negative or negligible speed-up should never be quantized regardless of its sensitivity score, as it degrades latency while injecting unnecessary quantization noise. This motivates use of per-layer latency profiling as a first-class signal in the quantization policy, rather than treating all layers as equally beneficial candidates for INT8 or relying strictly on theoretical bit allocation. We treat this implication as a hypothesis rather than an assumption, and evaluate it as an explicit ablation in Section~\ref{sec:fragmentation}.

\section{The RAMP Method}
\label{sec:ramp}
To address the MPQ problem, we propose RAMP (\textbf{R}obust \textbf{A}daptive \textbf{M}ixed-\textbf{P}recision), a robust and solver-free method. The RAMP pipeline operates through three stages:

\begin{enumerate}
	\item \textbf{Sensitivity Profiling:} Leveraging the JSD metric to accurately identify layers vulnerable to quantization.
	\item \textbf{Distribution-Adaptive Thresholding:} Using K-Means clustering to autonomously discover natural sensitivity breakpoints within the layer distribution.
	\item \textbf{Candidate Policy Generation:} Deriving one policy per ordered cluster centroid, from which the operational policy is selected on the Pareto front.
\end{enumerate}

This end-to-end pipeline is formalized in Algorithm \ref{alg:ramp}. Compared to state-of-the-art MPQ methods that rely on computationally expensive search strategies, such as ILP or RL, our approach incurs minimal computational overhead while obviating the need for arbitrary, hand-tuned sensitivity thresholds.

\begin{algorithm}[h]
	\caption{RAMP: Robust Adaptive Mixed-Precision Quantization}
	\label{alg:ramp}
	\footnotesize
	\SetKwInOut{Input}{Input}
	\SetKwInOut{Output}{Output}
	
	\Input{Pre-trained model $\mathcal{M}$ with $L$ layers, calibration dataset $\mathcal{D}$, number of clusters $K$, precision set $B = \{b_{\text{INT8}}, b_{\text{FP32}}\}$}
	\Output{Set of candidate policies $\Pi = \{\pi_1, \pi_2, \dots, \pi_K\}$}
	
	\BlankLine
	\tcp{Step 1: Sensitivity Profiling via OAT}
	\ForEach{layer $l \in \{1, \dots, L\}$}{
		Quantize only layer $l$ to $b_{\text{INT8}}$, keeping others at $b_{\text{FP32}}$\;
		Compute quantized output logits $\hat{f}_l$ and FP32 reference logits $f_l$ using $\mathcal{D}$\;
		Calculate sensitivity $\Omega_l = M(f_l, \hat{f}_l)$ (Eq. \ref{eq:sensitivity})\;
	}
	Construct the sensitivity set $\Omega = \{\Omega_1, \Omega_2, \dots, \Omega_L\}$\;
	
	\BlankLine
	\tcp{Step 2: Distribution-Adaptive Thresholding}
	
	Partition $\Omega$ into $K$ clusters via 1D K-Means $\rightarrow \{c_1, \dots, c_K\}$ (Eq.~\ref{eq:kmeans})\;
	Sort cluster centroids: $\mu_{(1)} < \mu_{(2)} < \dots < \mu_{(K)}$\;
	
	\BlankLine
	\tcp{Step 3: Candidate Policy Generation}
	\For{$k \in \{1, \dots, K\}$}{
		Define policy $\pi_k$ as the assignment $P = \{p_1, \dots, p_L\}$ where
		$p_l = \begin{cases} b_{\text{INT8}} & \text{if } \Omega_l \leq \mu_{(k)} \\ b_{\text{FP32}} & \text{otherwise} \end{cases}$\;
	}
	
	\Return $\Pi = \{\pi_1, \pi_2, \dots, \pi_K\}$ for Pareto elbow selection\;
\end{algorithm}

To determine the optimal operational policy that best balances accuracy preservation and latency reduction, we construct a Pareto front by plotting the accuracy versus latency for the $K$ candidate policies generated by RAMP. The optimal policy is selected using the knee-point (elbow) method, identifying the point of maximum curvature, which represents the optimal trade-off where additional latency reductions would necessitate a disproportionate sacrifice in model accuracy.

We conduct an end-to-end validation of our method using the models and dataset introduced in Section~\ref{sec:mpq}, alongside the hardware and framework specified in Section~\ref{sec:methodology_hardware}.

\subsection{End-to-End Validation}
Based on the preceding sensitivity and hardware profiling analyses, we evaluate our distribution-adaptive K-Means thresholding method using three hardware-agnostic sensitivity criteria, corresponding to the best-performing representative of each proxy family identified in Section~\ref{sec:mpq}: output-space (JSD), parameter-space (STD) and gradient-space (HAWQ-V2). A fourth, hardware-aware variant (JSD+Latency) is introduced and analysed as an ablation in Section~\ref{sec:fragmentation}. This evaluation is conducted across both target CPU platforms. As reference points we include the FP32 model and uniform INT8 quantization, the fixed PTQ scheme described in Section~\ref{sec:background}, whose end-to-end speed-up was already reported in Figure~\ref{fig:speed-up-end-to-end}. We consider two metrics: test set accuracy and median inference time per image (latency). The optimal policy is selected using the knee-point method described before.

Table~\ref{tab:main_results} summarizes the empirical results of this end-to-end validation. \textit{First}, uniform INT8 collapses on three of the four architectures, losing between $28$ and $70$ accuracy points, and only the classical residual CNN survives it. This underscores the necessity of mixed-precision policies. \textit{Second}, some hardware-agnostic baselines fail inconsistently depending on the network topology: STD collapses in 2 of the 8 configurations and HAWQ-V2 in 4, and their failure sets are disjoint, so neither heuristic dominates the other. Their failures concentrate on architectures with highly skewed, non-residual sensitivity distributions, where second-order weight approximations misrank the critical layers. Measuring sensitivity directly in the output-activation space via JSD avoids this failure mode, as it observes the end-to-end effect of the perturbation rather than a local curvature estimate.

\begin{table*}[h]
	\centering
	\resizebox{\textwidth}{!}{%
		\setlength{\tabcolsep}{3pt}
		\begin{tabular}{l cc cc cc cc || cc cc cc cc}
			\toprule
			& \multicolumn{8}{c||}{\textbf{Apple M1}} & \multicolumn{8}{c}{\textbf{Raspberry Pi 5 (Cortex-A76)}} \\
			\cmidrule(lr){2-9} \cmidrule(lr){10-17}
			& \multicolumn{2}{c}{ResNet-18} & \multicolumn{2}{c}{EffNet-B0} & \multicolumn{2}{c}{ConvNeXt-T} & \multicolumn{2}{c||}{TinyViT} & \multicolumn{2}{c}{ResNet-18} & \multicolumn{2}{c}{EffNet-B0} & \multicolumn{2}{c}{ConvNeXt-T} & \multicolumn{2}{c}{TinyViT} \\
			\cmidrule(lr){2-3} \cmidrule(lr){4-5} \cmidrule(lr){6-7} \cmidrule(lr){8-9}
			\cmidrule(lr){10-11} \cmidrule(lr){12-13} \cmidrule(lr){14-15} \cmidrule(lr){16-17}
			\textbf{Method} & Acc & Lat & Acc & Lat & Acc & Lat & Acc & Lat & Acc & Lat & Acc & Lat & Acc & Lat & Acc & Lat \\
			\midrule
			FP32 & 62.5 & 15.8 & 96.6 & 9.1 & 97.2 & 41.5 & 97.5 & 39.0 & 62.5 & 62.0 & 96.6 & 36.5 & 97.2 & 167.9 & 97.5 & 167.8 \\
			\midrule
			INT8 unif. & 61.5 & 2.9 & \clp{29.6} & 4.7 & \clp{55.0} & 23.7 & \clp{63.2} & 27.4 & 61.3 & 12.2
			& \clp{26.9} & 18.4 & \clp{33.1} & 90.4 & \clp{69.0} & 104.4 \\
			STD & 61.7 & 5.2 & 96.6 & 8.6 & \clp{79.7} & 20.1 & 89.6 & 27.1 & 61.6 & 25.9 & 96.6 & 35.2 & \clp{79.0} & 87.4 & 92.7 & 122.5 \\
			HAWQ-V2 & 61.7 & 4.8 & \clp{47.0} & 7.4 & 97.0 & 24.8 & \clp{82.1} & 27.0 & 62.0 & 20.7 & \clp{34.1} & 31.9 & 97.0 & 100.6 & \clp{80.8} & 122.5 \\
			\textbf{RAMP (JSD)} & 62.0 & 6.2 & 95.1 & 7.0 & 96.3 & 21.8 & 97.3 & 27.4 & 62.0 & 20.8 & 95.6 & 30.9 & 97.2 & 94.6 & 97.4 & 124.8 \\
			\midrule
			\multicolumn{17}{l}{\emph{Ablation: per-layer hardware filtering (Section~\ref{sec:fragmentation})}} \\
			\quad JSD + Latency & 62.1 & 5.5 & 95.0 & 7.0 & 97.2 & 22.4 & 97.3 & 27.4 & 62.5 & 49.7 & 95.3 & 30.2 & 97.2 & 115.5 & 97.4 & 144.1 \\
			\bottomrule
		\end{tabular}
	}
	\caption{End-to-end validation. Results report absolute Top-1 accuracy (\%) and inference latency (ms). Shaded cells mark collapse, as defined in Section~\ref{sec:mpq}.}
	\label{tab:main_results}
\end{table*}

Our proposed JSD-driven thresholding completely avoids performance degradation across all evaluated models and hardware platforms. For instance, on the highly sensitive TinyViT architecture, JSD achieves near-lossless accuracy (97.3\% on the Apple M1 and 97.4\% on the Raspberry Pi 5) compared to the FP32 baseline (97.5\%), while unlocking substantial hardware acceleration (e.g., reducing ConvNeXt-Tiny latency from 168~ms to 95~ms on the Raspberry Pi 5). Overall, end-to-end robustness follows the ranking established in Table~\ref{tab:recall_topk}: JSD (0.766 mean recall) incurs no failures, whereas HAWQ-V2 (0.509) and STD (0.496) fail in 4 and 2 of the 8 configurations respectively. Recall@Top-$N\%$ over the sensitivity proxies is therefore predictive of deployment robustness, which validates the metric selection of Section~\ref{sec:mpq}.

\subsection{Graph Fragmentation}
\label{sec:fragmentation}
We now evaluate the hypothesis formulated in Section~\ref{sec:methodology_hardware}, namely that layers yielding negligible speed-up should be excluded from quantization. We implement it as a streamlined two-stage pipeline, denoted JSD+Latency. \textit{First}, we filter out layers that do not yield a hardware speed-up greater than a given threshold $t$ (e.g., $t=1$). \textit{Second}, we apply the sensitivity-driven policy allocation of Section~\ref{sec:ramp} on the remaining layers. This strategy incorporates real hardware measurements rather than hardware simulations, using a simple and time-efficient procedure. To assess it, we compare JSD-only against JSD+Latency across all configurations. As shown in Table~\ref{tab:main_results}, both methods prevent accuracy collapse in all tested architectures (0 failures out of 8 configurations), with mean accuracy drops of $0.59\%$ and $0.45\%$ respectively, a negligible difference. The filter does not compensate for this on the latency side either: it degrades end-to-end inference time in 4 out of 8 configurations, improves it in 2, and leaves it unchanged in 2. Critically, these degradations are not uniformly distributed across platforms. On the Apple M1 the filter is neutral (at most $21.8$\,ms to $22.4$\,ms on ConvNeXt-Tiny), whereas on the Raspberry Pi~5 it degrades 3 out of 4 models: ConvNeXt-Tiny increases from $94.6$\,ms to $115.5$\,ms and TinyViT from $124.8$\,ms to $144.1$\,ms, both at identical accuracy, while ResNet-18 increases from $20.8$\,ms to $49.7$\,ms ($2.4\times$) for a gain of $0.5$ accuracy points.

\begin{figure}[h]
	\centering
	\includegraphics[width=0.9\linewidth]{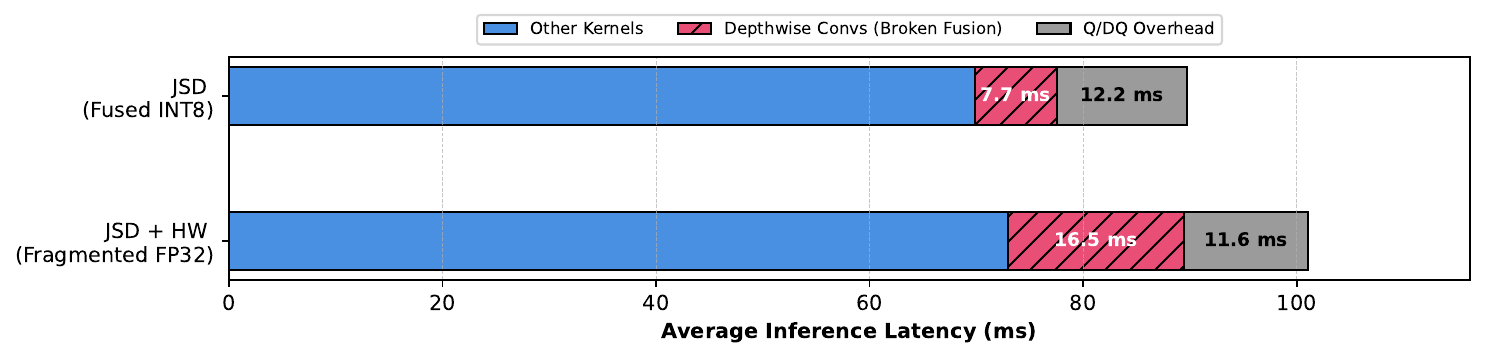}
	\caption{\textbf{Impact of Graph Fragmentation on ConvNeXt-Tiny (Cortex-A76).} The unoptimized, isolated FP32 fallback more than doubles the execution time of memory-bound depthwise convolutions (from 7.7~ms to 16.5~ms per inference), heavily degrading overall latency despite no significant increase in Q/DQ overhead.}
	\label{fig:graph_fragmentation}
\end{figure}

The mechanism behind this degradation is the execution engine rather than the individual operators. As shown in Figure~\ref{fig:graph_fragmentation}, forcing a subset of memory-bound layers into FP32 fragments contiguous INT8 subgraphs, introducing Quantize/Dequantize (Q/DQ) conversion nodes that prevent operator fusion and increase subgraph latency from $7.7$\,ms to $16.5$\,ms ($2.1\times$), a graph fragmentation effect documented in GPU inference compilers~\cite{tensorrt_qdq}. The cost of breaking a contiguous integer region therefore exceeds the latency saved by protecting the individual layers inside it, which explains why a filter that is provably correct at the layer level becomes counterproductive end-to-end.

This result invalidates the assumption that execution time is strictly additive across layers, on which the ILP formulations used by criterion-based MPQ methods~\cite{yao2021hawqv3, akbulut2026infoq} directly rely: a per-layer cost table cannot express a penalty that depends on the precision assigned to neighbouring layers. It also refines the hardware heterogeneity characterized in Section~\ref{sec:methodology_hardware}. Hardware still dictates the outcome, as evidenced by the fact that the filter is harmful on the Cortex-A76 and neutral on the Apple M1, but it is not actionable as a per-layer gating signal, since the effective unit of hardware awareness on CPU execution engines is the contiguous integer subgraph. Consequently, we exclude latency profiling from RAMP: on ARM64 platforms with
native INT8 SIMD support, JSD alone selects a low-sensitivity layer pool that provides adequate end-to-end speed-up without any on-device measurement, and the resulting policy transfers unchanged across both platforms.

\section{Conclusion and Future Work}
\subsection{Conclusion}
We conducted a comprehensive analysis of sensitivity metrics across four distinct model architectures and two ARM64 CPU platforms, demonstrating that information-theoretic metrics based on final output logits most accurately capture model degradation. Specifically, our empirical study shows that JSD, coupled with 1D K-Means clustering, consistently prevents the catastrophic accuracy collapse observed in both uniform quantization and state-of-the-art Hessian-based methods, particularly in modern Vision Transformers. Consequently, RAMP achieves near-lossless accuracy while delivering substantial inference speed-ups across diverse ARM64 platforms.

Crucially, we characterize the phenomenon of graph fragmentation in CPU execution engines, revealing that maintaining contiguous integer subgraphs has a more significant impact on end-to-end latency than the optimization of isolated layer-wise execution times. These findings challenge the prevailing assumption that mixed-precision latency can be accurately modeled as a simple additive sum of independent layer costs. Ultimately, RAMP offers a robust and practical pipeline for deploying state-of-the-art vision models on resource-constrained edge devices, eliminating the need for retraining or architecture-specific tuning.

\subsection{Future Work}
Four limitations delimit the scope of our results and define its natural extensions. \textit{First}, while our distribution-adaptive thresholding effectively generates robust MPQ policies, the realization of physical speed-ups remains inherently tied to the underlying execution engine: preliminary evaluations on x86-64 (Intel Core i9-12900KF) using the default ONNX Runtime CPU provider resulted in unexpected latency degradation across all models, and achieving the theoretical hardware efficiency strictly necessitates vendor-optimized backends such as OpenVINO, whose integration for non-ARM architectures remains a key avenue for future work. \textit{Second}, the proposed pipeline has been exclusively evaluated on image classification benchmarks; dense prediction tasks and natural language processing inherently exhibit fundamentally different activation distributions and structural vulnerabilities, so validating the transferability of our method across these domains remains critical. \textit{Third}, as a data-driven PTQ method, the reliability of JSD relies heavily on the representativeness of the calibration dataset: if it fails to capture out-of-distribution features or rare edge cases, the K-Means algorithm may misclassify structurally critical layers as low-sensitivity, making calibration-free (zero-shot) thresholding an open challenge. \textit{Finally}, our one-at-a-time profiling does not take into account interactions between layers; while the computed ranking is sufficient to generate robust policies in the binary FP32/INT8 regime, the resulting policies may be suboptimal, and extending the sensitivity analysis to block-wise measurement is a natural next step.

\section*{Acknowledgments}
This work was partially funded by two projects: (1) DARE SGA1, which has received funding from the European High-Performance Computing Joint Undertaking (JU) under grant agreement No 101202459; and (2) ODISSEE, which has received funding from the European Union's Horizon Europe research and innovation programme under grant agreement No 101188332.

\bibliographystyle{ieeetr}
\bibliography{references}

\end{document}